\documentclass{article}
\usepackage{tccml_iclr2025_conference, times}
\usepackage[style=apa]{biblatex}
\usepackage{amsfonts}
\usepackage{amsmath}
\usepackage{graphicx} 
\usepackage{svg}

\title{Do machine learning climate models work in changing climate dynamics?}

\author{Maria Conchita Agana Navarro \\
Centre for Artificial Intelligence, Department of Computer Science\\
University College London\\
\texttt{m.navarro.23@ucl.ac.uk} \\
\And
Geng Li \\
The Hong Kong University of Science and Technology \\
  \texttt{glibd@connect.ust.hk} \\
\AND
Theo Wolf \\
University of Oxford \\
\texttt{theo@robots.ox.ac.uk}
\AND
María Pérez-Ortiz \\
Centre for Artificial Intelligence, Department of Computer Science\\
University College London\\
\texttt{maria.perez@ucl.ac.uk} \\
}

\date{January 2025}

\begin{document}

\maketitle
\begin{abstract}
Climate change is accelerating the frequency and severity of unprecedented events, deviating from established patterns. Predicting these out-of-distribution (OOD) events is critical for assessing risks and guiding climate adaptation. While machine learning (ML) models have shown promise in providing precise, high-speed climate predictions, their ability to generalize under distribution shifts remains a significant limitation that has been underexplored in climate contexts. This research systematically evaluates state-of-the-art ML-based climate models in diverse OOD scenarios by adapting established OOD evaluation methodologies to climate data. Experiments on large-scale datasets reveal notable performance variability across scenarios, shedding light on the strengths and limitations of current models. These findings underscore the importance of robust evaluation frameworks and provide actionable insights to guide the reliable application of ML for climate risk forecasting.
\end{abstract}

\section{Introduction}

Climate change is driving more frequent and unprecedented extreme climate events \parencite{kornhuber_global_2024, IPCC2023, RN4}. Climate models are vital tools for managing these risks, simulating interactions among the atmosphere, oceans, land, and ice \parencite{climateset-doc}. Machine learning (ML) has emerged as a transformative addition to climate modeling, offering computationally efficient approximations of climate processes \parencite{kaltenborn2023climatesetlargescaleclimatemodel}. However, ML-based models have been shown to fail to generalize well under distribution shifts, when the real-world data deviates significantly from the ML training data \parencite[]{woods}. This raises concerns about the reliability of using ML for climate modelling in out-of-distribution (OOD) scenarios. For example, unprecedented combinations of greenhouse gas concentrations or tipping points such as ice sheet collapse could challenge ML models' predictive robustness. While benchmarks like ClimateBench \parencite{https://doi.org/10.1029/2021MS002954} and ClimateSet \parencite{kaltenborn2023climatesetlargescaleclimatemodel} provide standardized ML-ready datasets, they currently evaluate models under limited conditions (e.g. only SSP-2.45). Even recent, state-of-the-art models like NeuralGCM \parencite{Kochkov_2024} face challenges when extrapolating to extreme warming scenarios. Understanding the reliability of climate variable predictions, such as temperature and precipitation, is critical, with implications for early warning systems, agricultural planning, and water resource management. In high-vulnerability regions, where adaptive capacity is limited, unreliable predictions can exacerbate risks.

This paper introduces a novel evaluation methodology that integrates principles of OOD testing to assess the robustness of ML-based climate models under shifting climate dynamics, addressing a gap in more holistic assessments of model robustness. This study contributes to understanding the reliability of ML models in predicting climate risks under real-world conditions.

\section{Data and Methodology}

Climate data is time-series data, consisting of sequences of data points indexed over time; climate observations, such as temperature or precipitation, are collected at different time steps. Each data point \( X_t \) represents a specific observation at time \( t \in T \) and the corresponding label \( Y_t \) could represent a future climate state, such as temperature or precipitation projections. 

The WOODS framework \parencite[]{woods} provides a systematic approach to evaluating OOD performance for time-series tasks across diverse domains, including neurophysiology, sign language, and energy consumption. OOD settings occur when ML models trained on one type of data distribution encounters data from a different, unseen distribution. The WOODS framework proposes different types of shifts in time-series data. Below we outline how these shifts can be applied to ML climate modeling tasks.

\subsection{Adapting Existing OOD Evaluation Frameworks to ML Climate Models}

Climate data is typically drawn from different geographical regions or climate regimes. We can define each domain \( d \) as a specific geographical area or climate regime, with associated distribution \( P_d(X_t, Y_t) \). For ML climate models, the objective is to create a model \( f \) that generalizes well across different climate domains. To achieve this, we define the training data as:

\[
E_{\text{train}} = \{ d_1, d_2, ..., d_n \}
\]

where \( E_{\text{train}} \) represents the set of geographical regions or climate regimes used for training. The model is then evaluated on unseen domains \( E_{\text{all}} \), where \( E_{\text{train}} \subseteq E_{\text{all}} \).

To quantify generalization, we minimize the model's worst-case risk across all possible domains:

\[
\min_f \max_{d \in E_{\text{all}}} R_d(f)
\]

where \( R_d(f) \) is the risk or loss on domain \( d \), defined as:

\[
R_d(f) = \mathbb{E}_{(X_t, Y_t) \sim P_d(X, Y)} [\ell(f(X_t), Y_t)]
\]

where \( \ell \) is the loss function, and \( f(X_t) \) is the predicted climate state.

WOODS highlights two key scenarios for distribution shifts in time-series data:
\begin{enumerate}
    \item \textbf{Time-Domain Shifts:} Occur when the data distribution changes over time, which can be due to long-term trends, seasonal variations, or unexpected events.
    \item \textbf{Source-Domain Shifts:} Occur when training data comes from a domain that differs in underlying factors or conditions, such as different data sources.
\end{enumerate}

ML climate models are susceptible to both time and source-domain shifts. Climate data distributions can change over time, due to seasonal patterns, long-term climate trends like global warming, or events such as El Niño. Human-driven changes, such as land-use changes and urbanization, may also contribute to temporal shifts in climate data. Climate models can also encounter domain-shifts with training data originating from  regions or climate regimes that differ from data encountered in test settings or real-world use. Examples include shifts based on differences in the datasets' geographical regions (e.g., tropical vs. temperate climates) or data sources (e.g., climate change scenarios defined by Shared Socioeconomic Pathways vs ground-based observations). 

Building on time-domain and source-domain shifts, we propose two distinct methodologies to evaluate the ability of ML climate models to generalize under such conditions.

\textbf{Method 1 - Split Based on Time Period (Time-Domain Shift):} This method evaluates how well ML climate models generalize across data from different time periods. The models are trained on data from 1850–2014 and tested on data from 2015–2023, treating the recent period as an OOD scenario. We hypothesize that temporal shifts in climate patterns over time, such as increased global warming, may challenge the models' ability to generalize. This period is also chosen for testing, as ClimaX's pretraining data is limited to up to 2015 \parencite{nguyen2023climaxfoundationmodelweather}. 

\textbf{Method 2 - Evaluate Across Multiple SSP Scenarios (Source-Domain Shift):} This method assesses model performance when trained and tested on different Shared Socioeconomic Pathways (SSPs). Each SSP represents a distinct trajectory of global development, characterized by differences in socioeconomic factors such as population growth or energy use, which leads to varying climate forcing trajectories and temperature projections \parencite{climateset-doc}. We hypothesize that training on one set of SSPs and testing on another can simulate OOD conditions because the scenarios embody shifts in the distributions of climate drivers and their downstream effects. For example, SSP scenarios with high emissions (e.g. SSP5) differ in climate forcings compared to low-emission pathways (e.g., SSP1). While existing benchmarks often evaluate models against a single SSP scenario, this method expands on that by evaluating performance across multiple SSPs.

\section{Experiments and Results}

\subsection{Experiment Setup: Dataset, models, task, and evaluation metrics}
We use the ClimateSet dataset, which integrates inputs and outputs of temperature and precipitation from Input4MIPs and CMIP6 across 36 traditional climate models \parencite{kaltenborn2023climatesetlargescaleclimatemodel}. The dataset focuses on four major climate forcing agents (CO\textsubscript{2}, CH\textsubscript{4}, BC, and SO\textsubscript{2}) and four SSPs. It provides a diverse set of temporal and spatial features across multiple geographic regions, enabling us to assess model performance under various distribution shift conditions. The ClimateSet authors have also documented benchmark performance for state-of-the-art (SoTA) ML climate models, serving as a comparison point for the performance of these models in out-of-distribution settings. We utilize the implementation of these ML models within ClimateSet for consistency.
  
\textbf{Models:} We employ four advanced machine learning models (U-Net, ConvLSTM, ClimaX, and ClimaX Frozen) which have demonstrated strong performance on ClimateSet’s benchmark datasets.

\textbf{Task:} The core task being evaluated is climate emulation, where the ML models aim to replicate the outputs of traditional climate models. We use 5 traditional climate models (AW1-CM-1-1-MR, EC-Earth3, FGOALS-f3-L, BCC-CSM2-MR, MPI-ESM1-2-HR) used in ClimateSet's baseline experiments, allowing for comparison with their experiments.
 
 \textbf{Evaluation:} We evaluate the model's performance using the latitude-longitude weighted root mean squared error (RMSE), the metric reported from ClimateSet \parencite{kaltenborn2023climatesetlargescaleclimatemodel}. Predictions of monthly surface air temperature and precipitation are assessed against outputs from traditional climate models.

\subsection{Experiment Steps}

\textbf{1) Run ClimateSet baselines:} We first run the baseline; ClimateSet single emulator specifications \parencite{kaltenborn2023climatesetlargescaleclimatemodel}, for each ML model, establishing a clear baseline for comparison. Each emulator receives as input the climate forcing emission fields of CO\textsubscript{2}, CH\textsubscript{4}, BC, and SO\textsubscript{2}, with the output variables being climate model predictions of temperature and precipitation. The baseline training dataset includes 165 years of historical data (1850-2014) as well as 86-years of climate predictions for 3 SSP scenarios (2015-2100). For training, the historical data, SSP1-2.6, SSP3-7.0, and SSP5-8.5 are utilized. A random 10\% of this data is withheld for validation, and the SSP2-4.5 scenario is used for testing. 

\textbf{2) Evaluate model performance under distribution shifts:} 
We then run experiments assessing model performance under distribution shifts. For Method 1 (time-domain shift), the training data is restricted to data from the period 1850-2014, while the test data covers 2015-2023. For Method 2 (source-domain shift), the train-test split is varied across the different Shared Socioeconomic Pathways (SSPs). Three scenarios are tested, each SSP scenario being used as the test set in turn: SSP1-2.6, SSP3-7.0, and SSP5-8.5.

\subsection{Results}

The results reveal that ML models exhibit different strengths and weaknesses depending on the nature of the OOD data. Under a time-domain shift, most models demonstrated improved or comparable performance to the baseline for both temperature and precipitation predictions (Table \ref{tab:percent_change_all}). Improvements were indicated by negative percent changes in RMSE scores, with ClimaX consistently achieving the lowest RMSE scores across all climate models for both variables (Appendix \ref{appendixC} Figure \ref{fig:llrmse_temporal}). This aligns with ClimaX’s strong performance in ClimateSet’s benchmark experiment, showcasing its robustness in generalization. The enhanced performance under time-domain shifts suggests that ClimaX’s architecture may better capture temporal dependencies in climate data. Simpler models like U-Net also exhibited notable improvements, indicating that even less complex architectures can generalize well in this setting. The overall trend suggests the models may be biased towards patterns resembling historical data, with stronger generalization to near-future conditions.

\begin{table}
  \centering 
    \caption{Percent change (\%) in RMSE values between baseline performance and performance under time-domain and source-domain shifts for each ML model single-emulating surface air temperature (TAS) and precipitation (PR) across five climate models.}
  \includegraphics[width=0.9\textwidth]{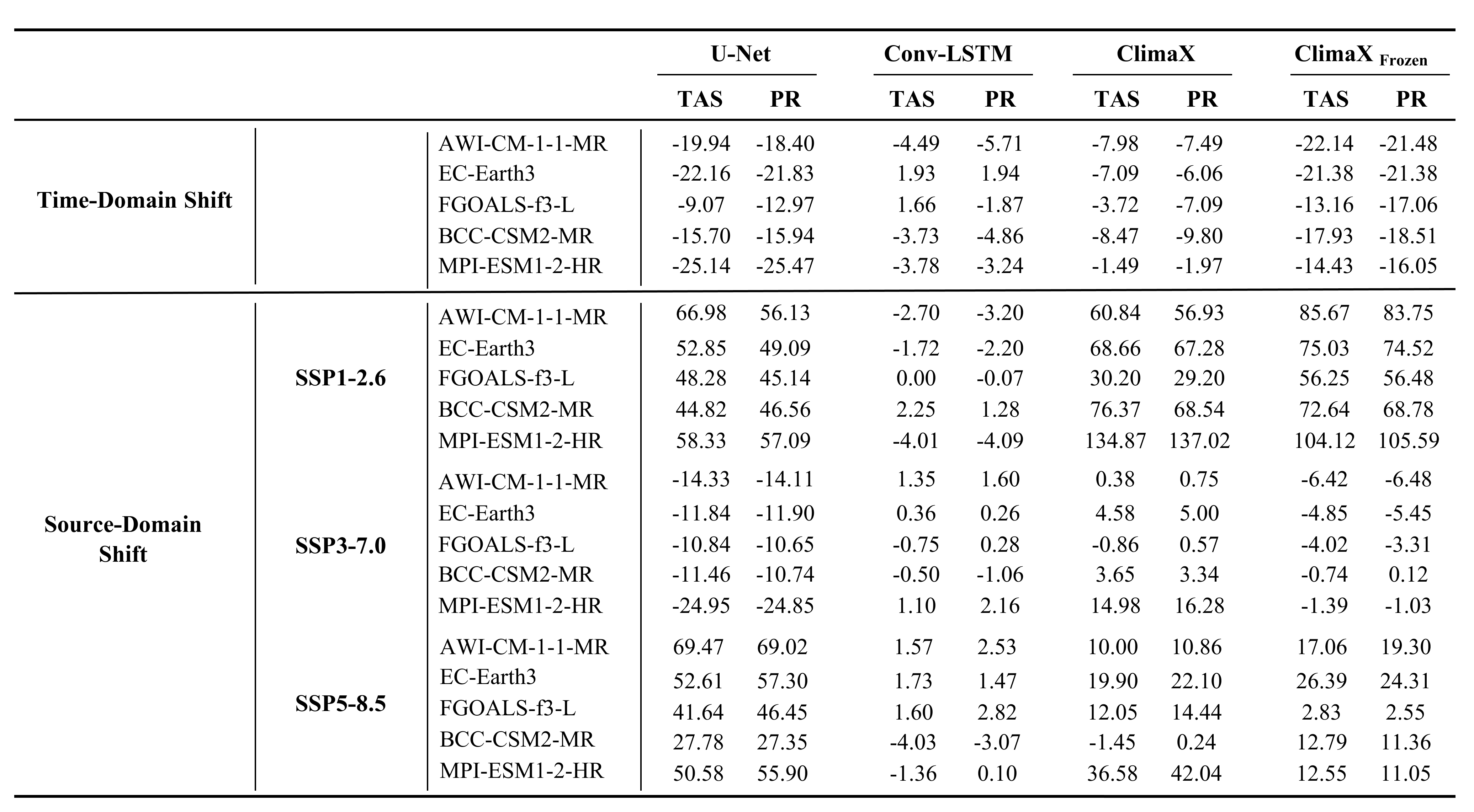}
  \label{tab:percent_change_all}
\end{table}


Under source-domain shifts, RMSE values generally increased compared to the baseline, indicating reduced performance (Table \ref{tab:percent_change_all}, Appendix \ref{appendixC} Figure \ref{fig:llrmse_domain}). However, certain scenarios presented exceptions where models performed comparably or better. Among the models, ConvLSTM emerged as the most consistent performer across scenarios, exhibiting minimal performance variation. It achieved the best results for SSP1-2.6, indicating strong adaptability to lower-emission scenarios. In contrast, U-Net struggled significantly with SSP1-2.6 and SSP5-8.5 but excelled in SSP3-7.0. ClimaX and ClimaX Frozen also encountered challenges with SSP1-2.6 and SSP5-8.5, suggesting potential limitations in capturing the variability introduced by extreme SSP scenarios.




Variations in performance carry relevance for policymakers and planners relying on climate forecasts; an observed 20–30\% increase in RMSE under specific OOD scenarios could, for example, indicate underestimations of the frequency and severity of temperature and precipitation events by ML models. Recognizing these inaccuracies during evaluation can inform decision-makers about models' predictive reliability in evolving climate conditions and guide any actions, such as further review, before decisions are made.

\section{Conclusion and Future Work}
This study evaluated the performance of SoTA ML  climate models under OOD scenarios, developing a novel evaluation framework tailored to performance assessment under evolving climate conditions.  Key recommendations include rotating time and source-domain scenarios between training and testing to ensure a comprehensive evaluation of model robustness across diverse climate domains. The framework is adaptable and can be used to evaluate other climate models. The exclusion of more recent SoTA ML models such as NeuralGCM \parencite[]{Kochkov_2024} limits the scope of comparison. Future research should aim to evaluate these SoTA models within the framework. Other promising avenues include expanding the framework to address other climate tasks beyond global climate model emulation, such as downscaling and regionalized predictions, as well as consider multiple random seeds and re-analysis datasets for more generalized insights. As ML's role advances to support our understanding, proactive evaluation will be crucial for reliable decision-making and ensuring these tools effectively contribute to addressing our changing climate.

\newpage
\printbibliography

\clearpage
\begin{appendix}

\section{Training Details}\label{appendixA}

Our baseline runs followed the ClimateSet single emulator specifications \parencite[]{kaltenborn2023climatesetlargescaleclimatemodel}:

\begin{itemize}

\item \textbf{Training Process:} Each emulator is trained on data from a single climate model, predicting outputs for an entire sequence of monthly data for each year. 

\item \textbf{Pre-Processing:}  The data has been pre-processed by ClimateSet to have a spatial resolution of approximately 250 km (144 x 96 longitude-latitude cells) and a temporal resolution of monthly data. The time series is divided into 1-year chunks, resulting in data with a shape of 
$\langle \textit{ scenarios, years * months, variables, longitude, latitude } \rangle$.

\item \textbf{Input and Output Shapes:} 
The input data has the shape $\langle \textit{ batch, sequence length, num vars, lon, lat }\rangle$, where the sequence length is 12 (monthly data). The output has the shape $\langle \textit{ batch, sequence length, 2, lon, lat }\rangle$, where the '2' corresponds to temperature (TAS) and precipitation (PR). 

\item \textbf{Training Parameters:} The models are trained for 50 epochs with an initial learning rate of 2e-4, using an exponential decay scheduler. For the non-frozen ClimaX models, training begins with a 5-epoch warm-up phase at 1e-8, followed by training at 5e-4.

\item \textbf{Loss:} The latitude-longitude weighted mean squared error (LLMSE) as implemented in \parencite[]{nguyen2023climaxfoundationmodelweather} is used.

\end{itemize}

\section{Evaluation Metrics Details}\label{appendixB}

The metric reported in our experiments is the Latitude-Longitude Weighted Root
Mean Squared Error (RMSE), implemented in \parencite[]{nguyen2023climaxfoundationmodelweather} and used in \parencite[]{kaltenborn2023climatesetlargescaleclimatemodel} ClimateSet experiments. It is a modified version of the Normalized Root Mean Squared Error (NRMSE) that accounts for varying grid sizes at different latitudes \parencite[]{nguyen2023climaxfoundationmodelweather}. 

\section{Detailed Results}\label{appendixC}

\begin{figure}[!htb]
\centering
\includegraphics[width=13cm]{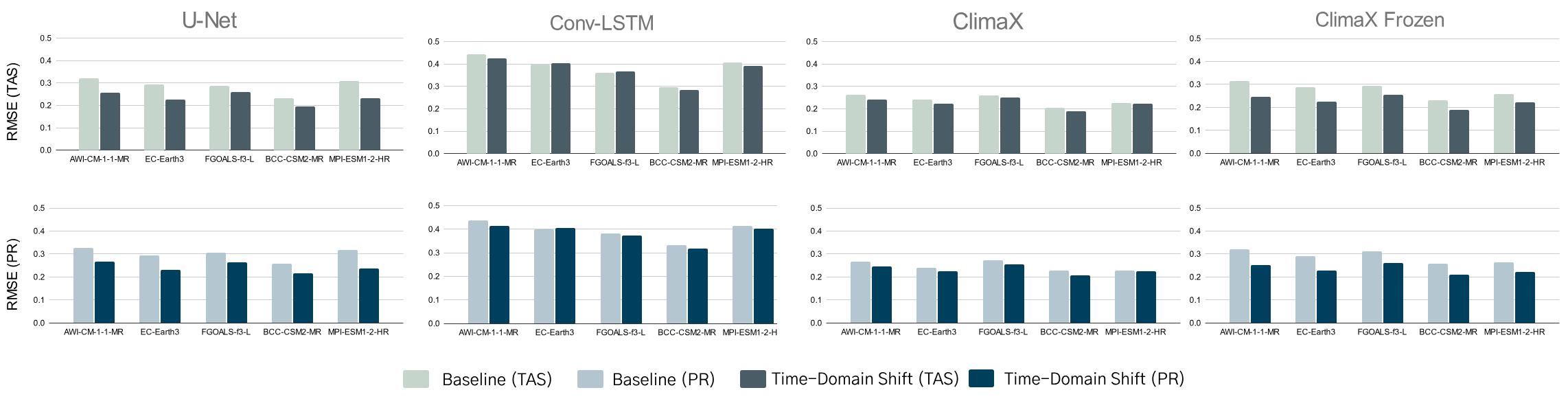}
\caption{Comparison of RMSE between baseline performance and performance under a time-domain shift for each ML model emulating surface air temperature (TAS) and precipitation (PR) across five climate models.}
\label{fig:llrmse_temporal}
\end{figure}

\begin{figure}[!htb]
\centering
\includegraphics[width=13cm]{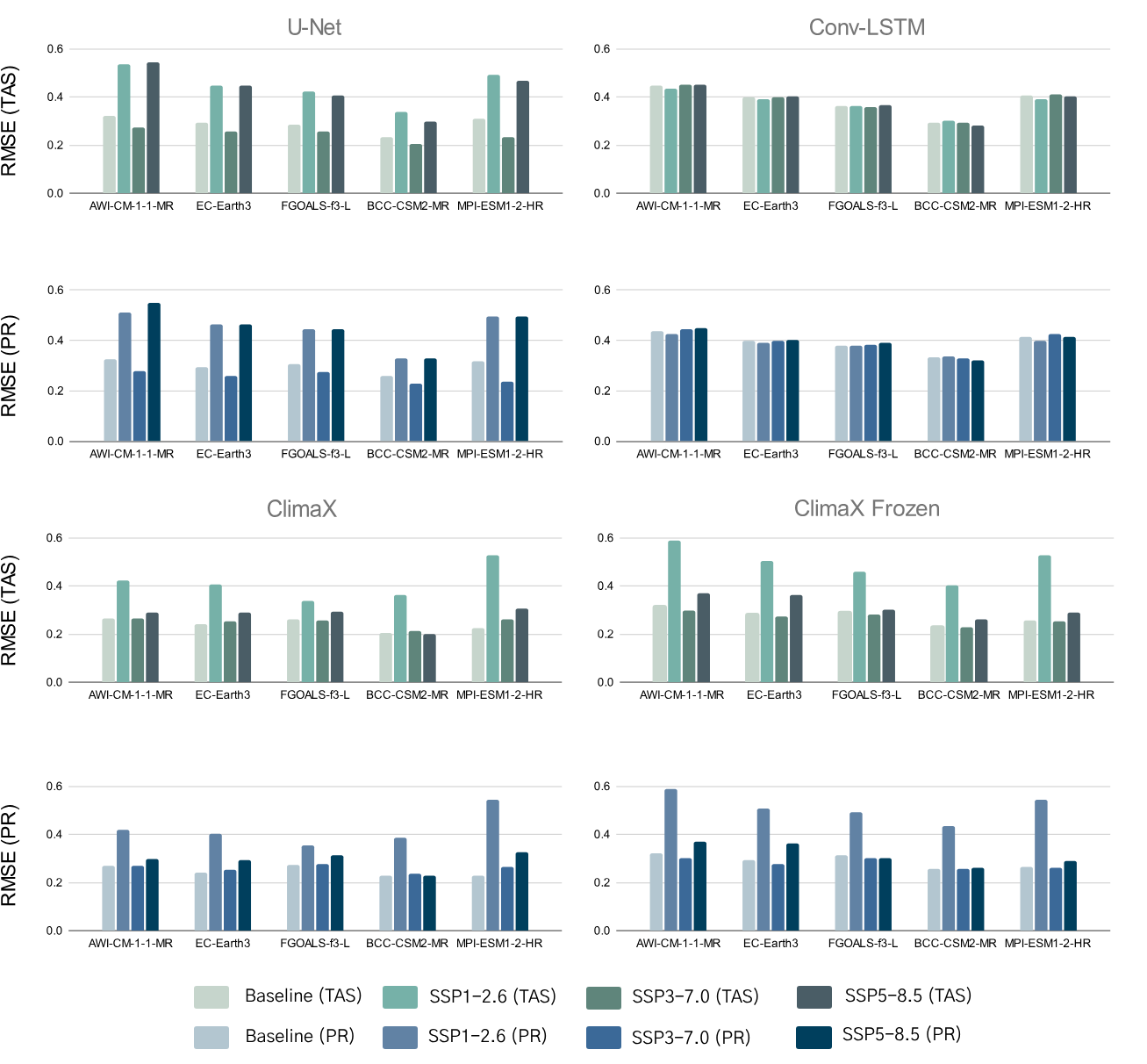}
\caption{Comparison of RMSE between baseline performance and performance under source-domain shifts for each ML model emulating surface air temperature (TAS) and precipitation (PR) across five climate models.}
\label{fig:llrmse_domain}
\end{figure}

Figures \ref{fig:llrmse_temporal} and \ref{fig:llrmse_domain} summarize the performance of all ML models for surface air temperature (TAS) and precipitation (PR). These comparisons evaluate baseline performance against performance under time-domain and source-domain distribution shifts.

\end{appendix}

\end{document}